\documentclass[letterpaper, 10 pt, conference]{ieeeconf}

\IEEEoverridecommandlockouts
\usepackage{amsmath,amssymb,amsfonts}
\usepackage{graphicx}
\usepackage{booktabs}
\usepackage{multirow}
\usepackage{xcolor}
\usepackage{hyperref}
\usepackage{algorithm}
\usepackage{algorithmic}
\usepackage{subcaption}
\usepackage{xspace}
\usepackage{bm}

\usepackage{wrapfig}
\usepackage{cite}

\hypersetup{colorlinks=true, linkcolor=blue, citecolor=blue, urlcolor=blue}

\usepackage[table]{xcolor}

\newcommand{\methodname}{\textsc{HiFlow}\xspace}

\newcommand{\R}{\mathbb{R}}

\title{\Large \bf \methodname: Tokenization-Free Scale-Wise Autoregressive\\
Policy Learning via Flow Matching}

\author{Daichi Yashima$^{*1,2}$, Koki Seno$^{*1}$, Shuhei Kurita$^{3,2}$, Yusuke Oda$^{2}$, Komei Sugiura$^{1}$%
\thanks{%
    $^{*}$Equal contribution.
    $^{1}$Keio University.
    $^{2}$National Institute of Informatics Research and Development Center for Large Language Models.
    $^{3}$National Institute of Informatics.
    {\tt\small ydaichi1207@keio.jp}
}%
\thanks{
    \small{
This work was partially supported by JSPS KAKENHI Grant Number 23K28168, JST Moonshot, and JST, CRONOS, Japan Grant Number JPMJCS24K6.
}
}
}

\begin{document}

\maketitle
\thispagestyle{empty}
\pagestyle{empty}

\begin{abstract}
  Coarse-to-fine autoregressive modeling has recently shown strong promise for visuomotor policy learning, combining the inference efficiency of autoregressive methods with the global trajectory coherence of diffusion-based policies.
  However, existing approaches rely on discrete action tokenizers that map continuous action sequences to codebook indices, a design inherited from image generation where learned compression is necessary for high-dimensional pixel data.
  We observe that robot actions are inherently low-dimensional continuous vectors, for which such tokenization introduces unnecessary quantization error and a multi-stage training pipeline.
  In this work, we propose Hierarchical Flow Policy (\methodname), a {tokenization-free} coarse-to-fine autoregressive policy
  that operates directly on raw continuous actions.
\methodname constructs multi-scale continuous action targets from each action chunk via simple temporal pooling.
Specifically, it averages contiguous action windows to produce coarse summaries that are refined at finer temporal resolutions.
The entire model is trained end-to-end in a single stage, eliminating the need for a separate tokenizer.
Experiments on MimicGen, RoboTwin 2.0, and real-world environments demonstrate that \methodname consistently outperforms existing methods including diffusion-based and tokenization-based autoregressive policies.
Our project page is available at \url{https://hiflow-6r47s.kinsta.page/}
\end{abstract}

\section{Introduction}
\label{sec:introduction}

Robotic manipulation is essential for autonomous robots operating in real-world environments.
A key challenge is learning visuomotor policies that map visual observations and proprioceptive states to action sequences, enabling robots to acquire manipulation skills from demonstrations and generalize across tasks and environments.
Given a dataset of expert trajectories, the generative modeling paradigm used for action prediction plays a central role in determining both policy performance and computational efficiency.

Autoregressive Modeling (AM) supports efficient single-pass inference and integrates naturally with sequence models~\cite{shafiullah2022bet,zhao2023act,chen2021decision}, but generating actions token-by-token limits its ability to capture global trajectory structure over longer horizons.
\begin{figure}[t]
  \centering
  \includegraphics[width=\linewidth]{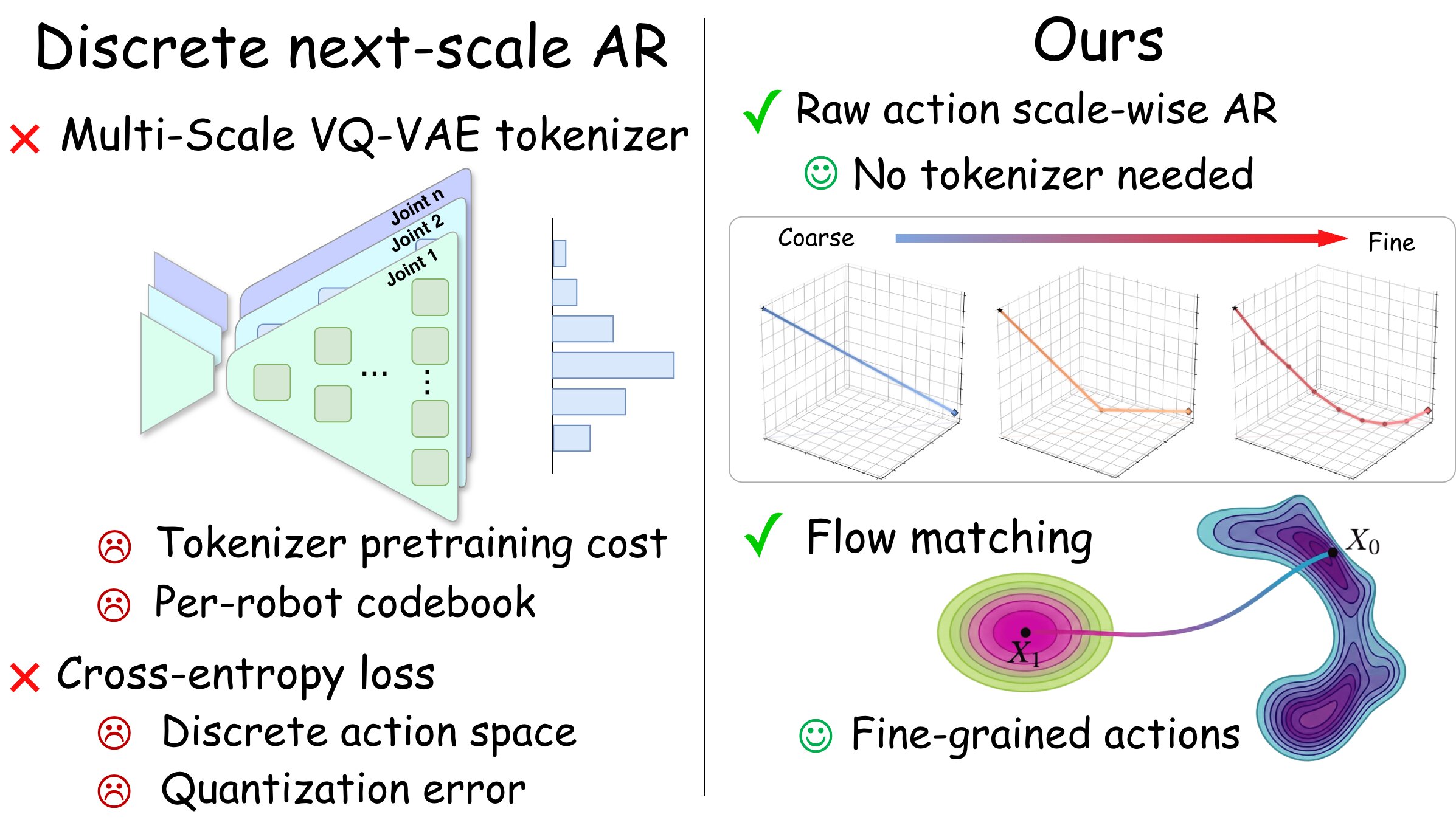}
  \caption{\textbf{Tokenized vs tokenization-free scale-wise policy learning.}
  Prior next-scale autoregressive policies discretize multi-scale action chunks using a VQ-VAE tokenizer and learn to predict code indices with cross-entropy, which introduces a separate tokenizer pretraining stage and incurs quantization error.
  \methodname instead refines raw continuous actions from coarse to fine temporal scales.
  A scale-wise autoregressive transformer provides coarse-to-fine conditioning, and a shared conditional flow matching module generates continuous actions at each scale, eliminating the tokenizer while preserving fine-grained control.}
  \label{fig:paradigm_comparison}
  \vspace{-2mm}
\end{figure}
Diffusion Modeling (DM) has recently achieved strong results in visuomotor policy learning by generating action sequences through iterative denoising~\cite{chi2023diffusionpolicy,ho2020ddpm}.
DM produces globally coherent trajectories, but the multi-step sampling process introduces substantial inference latency.

These two paradigms present complementary strengths: AM offers efficient inference but may lack global trajectory consistency, while DM provides strong global coherence at a higher computational cost.
To reconcile this trade-off, recent work introduces coarse-to-fine next-scale prediction for visuomotor policies~\cite{gong2024carp}.
Instead of generating actions strictly step by step, next-scale prediction produces action sequences from coarse global structure to fine-grained details, using a coarse-to-fine autoregressive decomposition in which each scale is conditioned on all coarser scales.
This paradigm builds on next-scale autoregressive generation originally proposed for image synthesis~\cite{tian2024var,ren2024flowar}, adapting the idea of progressively refining resolution from spatial to temporal scales.

Despite its promise, existing scale-wise visuomotor policies rely on discrete action tokenization, commonly implemented with VQ-VAE codebooks~\cite{vqvae}.
They first pretrain a tokenizer to map continuous actions to discrete indices at multiple temporal scales, then train an autoregressive predictor with cross-entropy over those indices.
This design couples policy quality to tokenizer quality, requires a multi-stage training pipeline, and introduces quantization error.

Building on the next-scale paradigm, we propose Hierarchical Flow Policy (\methodname), a coarse-to-fine policy that operates directly in continuous action space without discrete tokenization.
\methodname consists of two core components: a scale-wise autoregressive transformer (ScaleAR) that produces conditioning features at progressively finer temporal scales, and a shared flow matching network (ActionFlowNet) that generates continuous actions at each scale conditioned on these features.
We construct temporal scales through deterministic downsampling and model the per-scale conditional distributions using a scale-wise flow matching formulation.
As a result, the action trajectory is refined from a single-token global summary to the full-resolution action chunk, preserving the efficiency and global structure of next-scale prediction while eliminating the need for a learned tokenizer and a multi-stage training pipeline.

Figure~\ref{fig:paradigm_comparison} compares the two paradigms.
Tokenized scale-wise policies first pretrain a VQ-VAE to convert continuous actions into discrete code indices at each scale, then predict those indices with cross-entropy loss.
\methodname replaces the learned tokenizer with deterministic temporal averaging to construct multi-scale action targets, and employs conditional flow matching in place of cross-entropy to generate continuous actions at each scale.
This design removes the separate tokenizer pretraining stage and avoids the quantization error inherent in discrete codebooks, yielding a single-stage training pipeline that operates entirely in continuous action space and requires no embodiment-specific tokenizer design.

In summary, our contributions are:
\begin{itemize}
  \item We revisit discrete action tokenization in next-scale visuomotor policies and introduce a tokenizer-free formulation that eliminates the multi-stage training pipeline and reduces embodiment-specific overhead.
  \item We propose \methodname, a coarse-to-fine policy that unifies the next-scale autoregressive structure with scale-wise flow matching to model continuous action distributions.
  \item We conduct comprehensive experiments on MimicGen, RoboTwin 2.0, and real-world tasks, demonstrating that \methodname outperforms baseline methods.
\end{itemize}

\section{Related Work}
\label{sec:related_work}

\subsection{Visuomotor Policy Learning}

Behavior cloning (BC) learns policies by directly generating actions from observations~\cite{pomerleau1989alvinn,ravichandar2020recent}, but it suffers from compounding errors under distributional shift and has difficulty capturing complex action distributions.
Sequence modeling approaches reinterpret policy learning as next-token prediction, including Decision Transformer~\cite{chen2021decision}, Behavior Transformer~\cite{shafiullah2022bet}, and action chunking for improved temporal consistency~\cite{zhao2023act}.
Latent action representations further improve expressiveness~\cite{lee2024latentactions}, yet autoregressive token prediction often struggles to capture global trajectory structure.

Diffusion-based policies address these limitations by modeling action generation as conditional denoising.
Diffusion Policy~\cite{chi2023diffusionpolicy} demonstrates strong performance on a range of manipulation tasks, and subsequent work has improved its expressiveness through 3D representations~\cite{ze20243ddp}, sparse architectures~\cite{wang2024sdp}, hierarchical skills~\cite{liang2024skilldiffuser}, and consistency distillation for fewer-step inference~\cite{prasad2024consistency,lu2024manicm}.
Despite these advances, diffusion policies require multiple sequential denoising steps, introducing additional inference overhead.

Flow-based generative policies mitigate this overhead by learning continuous velocity fields that require fewer sampling steps.
Flow matching~\cite{lipman2023flow} and consistency models~\cite{song2023consistency} enable high-quality generation with fewer steps, and have been recently applied to robot control and vision-language-action models~\cite{pi0-black25rss,pi05-black25corl,flowpolicy-zhang-aaai25}.
While these methods improve sampling efficiency, they typically operate at a single temporal scale and do not explicitly model global-to-local trajectory refinement.

\subsection{Coarse-to-Fine Autoregressive Generation}

Scale-wise autoregressive modeling was introduced by VAR~\cite{tian2024var}, which replaces raster-scan token prediction with next-scale prediction across resolutions.
This paradigm has been extended to continuous generation, including FlowAR~\cite{ren2024flowar}, which integrates flow matching into multi-scale generation.

CARP~\cite{gong2024carp} adapts scale-wise autoregressive modeling to visuomotor policy learning by reinterpreting spatial scales as temporal scales over the action horizon.
Using multi-scale VQ-VAE tokenization~\cite{vqvae} and cross-entropy loss, CARP achieves strong performance with autoregressive efficiency.
However, discrete tokenization introduces quantization error, multi-stage training, and embodiment-specific tokenizers, all of which add overhead without clear benefit for low-dimensional continuous action spaces.
 These observations motivate a tokenizer-free approach that models multi-scale action distributions directly in continuous space.

\subsection{Flow Matching}

Diffusion models~\cite{ho2020ddpm} and flow matching~\cite{lipman2023flow} are generative frameworks that progressively transport a prior distribution toward the data distribution.
These frameworks have shown strong capabilities across images~\cite{sd3-esser24icml,peebles2023dit}, videos~\cite{pyramidal-jin25-iclr,moviegen-polyak24}, and robot action trajectories~\cite{pi0-black25rss,pi05-black25corl,flowpolicy-zhang-aaai25,maniflow-yan25corl,zhang2025reinflow,ze20243ddp,chi2023diffusionpolicy}.
However, they typically require many sampling steps during inference, limiting generation quality under a fixed computational budget~\cite{lipman2023flow,ho2020ddpm,sd3-esser24icml,sd-rombach22cvpr}.
To address this, several approaches incorporate autoregressive conditioning on previously generated outputs~\cite{ren2024flowar,tian2024var,ar-diffusion-sun25cvpr,ar-diffusion-wu23neurips,janusflow-ma25cvpr,infinitystar-liu25neurips,showo-xie25iclr,epona-zhang25iccv,transition-matching-shaul25neurips}.
This strategy enables high-quality generation while preserving the foundational formulation of the underlying generative frameworks~\cite{ren2024flowar,tian2024var}.
Our work builds on this continuous-space formulation and extends it to scale-wise action generation for visuomotor policies.

\begin{figure*}[t]
  \centering
  \includegraphics[width=\linewidth]{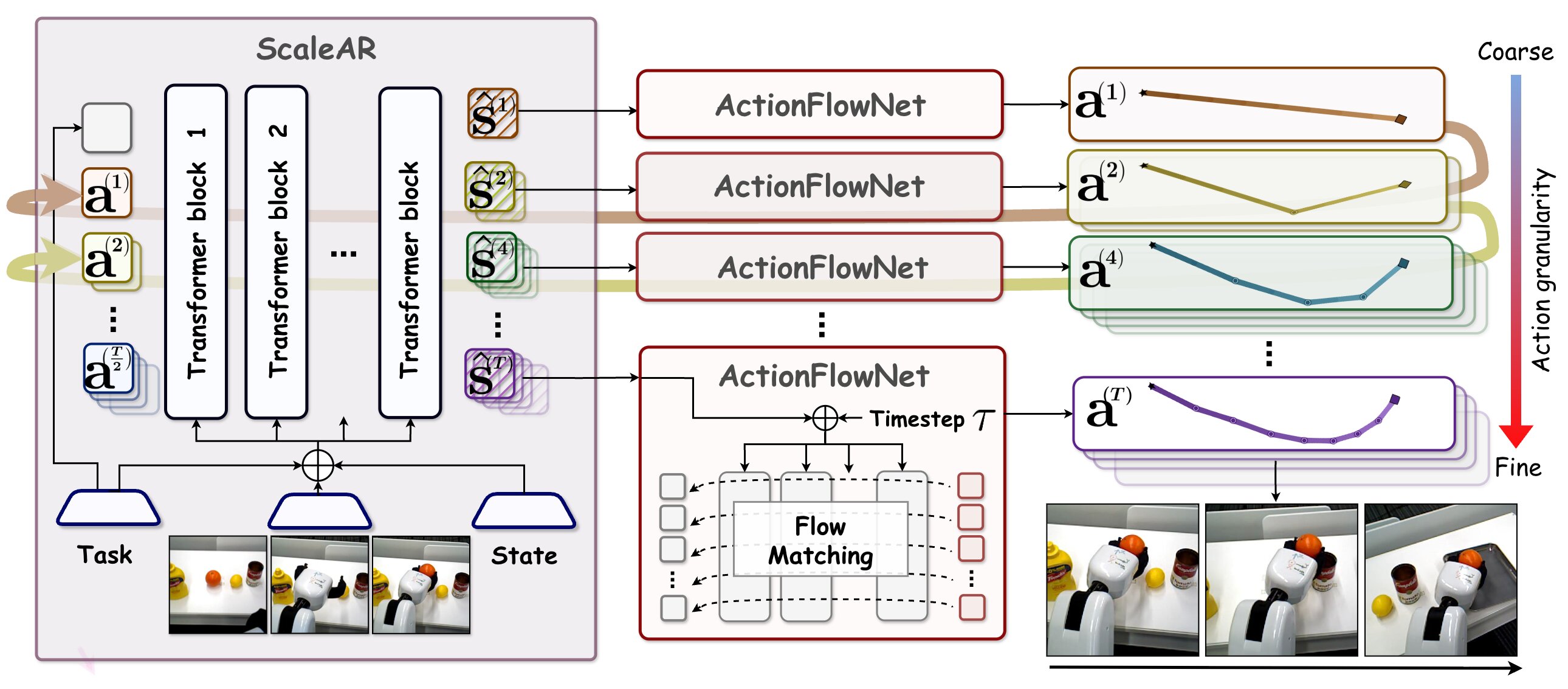}
  \vspace{2pt}
  \caption{
    \textbf{Architecture overview of \methodname.}
    Given visual observations, proprioceptive states, and a task identifier, the scale-wise autoregressive Transformer (ScaleAR) produces conditioning features at progressively finer temporal scales via a scale-wise causal mask.
    A shared ActionFlowNet then generates continuous actions at each scale through conditional flow matching, progressively refining the trajectory from a single-token global summary (scale~1) to the full $T$-step action chunk.
    The entire pipeline operates in continuous action space without any discrete tokenization.
  }
  \label{fig:architecture}
\end{figure*}

\section{Problem Statement}
\label{sec:problem_statement}
We address the problem of learning a multi-task visuomotor policy from a fixed dataset of expert demonstrations.
Following prior work~\cite{chi2023diffusionpolicy,gong2024carp}, we formulate this as an action sequence prediction problem under the behavior cloning framework, where the objective is to learn a policy that predicts future action sequences conditioned on observations.
At each timestep, the robot receives a visual observation consisting of multiple RGB images, along with a proprioceptive state encoding the robot's joint configuration or end-effector pose.
In the multi-task setting, a task identifier is also provided to condition the policy on the desired task.
These inputs together constitute the full observation.
The policy outputs a sequence of consecutive actions over a short temporal horizon, where each action is a continuous vector specifying the desired robot command (e.g., end-effector displacement and gripper aperture).

\section{Method}
\label{sec:method}

We present \methodname, a tokenization-free coarse-to-fine autoregressive visuomotor policy that generates action chunks via scale-wise flow matching.
We first review the scale-wise autoregressive paradigm (Sec.~\ref{subsec:background}), then describe how \methodname removes the need for discrete tokenization.

\subsection{Preliminary: Scale-Wise Autoregressive Generation}
\label{subsec:background}

Standard autoregressive models predict actions one token at a time in a fixed sequential order~\cite{shafiullah2022bet,chen2021decision}.
In image generation, VAR~\cite{tian2024var} and FlowAR~\cite{ren2024flowar} proposed an alternative: scale-wise autoregressive generation.
Instead of a fixed sequential order, this approach generates the output at multiple resolutions, progressing from coarse to fine.
At the coarsest scale, a small number of tokens summarize the global structure; at each subsequent scale, the number of tokens increases, progressively adding finer detail until the full-resolution output is recovered.
A scale-wise causal structure ensures that each scale can only depend on coarser scales, preserving the autoregressive factorization while allowing all tokens within a scale to be generated in parallel.

The adaptation of this paradigm to visuomotor policy learning~\cite{gong2024carp} inherited the discrete tokenization design from image generation: a VQ-VAE~\cite{vqvae} is first pretrained to learn a finite codebook that maps continuous data to discrete indices at each scale, and an autoregressive Transformer then predicts the next scale's indices via cross-entropy classification over the codebook vocabulary.
This two-stage pipeline, while effective for high-dimensional pixel data, requires careful tokenizer design and hyperparameter tuning, and introduces quantization error that propagates through the generation pipeline.

\subsection{\methodname Overview}
\label{subsec:overview}

\methodname replaces discrete tokenization with a fully continuous formulation for scale-wise autoregressive action generation.
The key observation is that robot actions are low-dimensional continuous vectors (typically 7--14 DoF), fundamentally different from the high-dimensional pixel data for which VQ-VAE tokenization was designed.
This makes discrete tokenization unnecessary and allows operating directly in continuous action space, avoiding the limitations described above.

As shown in Figure~\ref{fig:architecture}, \methodname constructs multi-scale action representations through simple temporal averaging, a parameter-free operation that stays in the original continuous action space (Sec.~\ref{subsec:multiscale}).
A scale-wise autoregressive Transformer (ScaleAR) then generates conditioning features at each temporal scale following the coarse-to-fine dependency structure (Sec.~\ref{subsec:ar_transformer}).
Instead of generating discrete codebook indices via cross-entropy, a shared lightweight flow matching module (ActionFlowNet) generates continuous actions at each scale by learning a velocity field that transports Gaussian noise to the target action distribution (Sec.~\ref{subsec:flow_matching}).
The entire model is trained end-to-end in a single stage, eliminating the need for a pretrained tokenizer (Sec.~\ref{subsec:training_inference}).

\subsection{Tokenization-Free Multi-Scale Action Representation}
\label{subsec:multiscale}

We construct multi-scale action representations by partitioning the action chunk along the temporal axis at progressively finer granularities and averaging each group.
Formally, given a ground-truth action chunk $\mathbf{a} \in \R^{T \times D}$ with $T$ timesteps and $D$-dimensional actions, we construct multi-scale targets by temporal averaging.
At each scale $i \in \mathcal{S} = \{1, 2, 4, \ldots, T\}$, consecutive groups of $T/i$ actions are averaged to generate $i$ representative actions:
\begin{align}
  \label{eq:multiscale}
  \mathbf{a}^{(i)} &= \text{Down}(\mathbf{a}, i) \in \R^{i \times D},
\end{align}
where $\text{Down}(\cdot, \cdot)$ partitions the $T$ timesteps into $i$ equal groups and returns the mean action of each group.

The coarsest scale ($i{=}1$) captures the global trajectory centroid, while finer scales progressively resolve temporal details.
Because the representation remains in the original continuous action space, it preserves Euclidean geometry and enables smooth gradient flow through subsequent flow matching.
This entire decomposition is parameter-free and adds negligible compute, in contrast to learned tokenizers that require a separate training stage~\cite{gong2024carp}.

\subsection{Scale-Wise Autoregressive Transformer}
\label{subsec:ar_transformer}

The scale-wise autoregressive Transformer (ScaleAR) produces conditioning features for each temporal scale, following the coarse-to-fine dependency structure.
Visual observations are encoded by a shared vision backbone (e.g., ResNet), concatenated with task and proprioceptive features, and fused through an MLP into a global condition vector $\mathbf{c}_\text{global} \in \R^{D}$.
This vector modulates all Transformer blocks via Adaptive Layer Normalization (AdaLN)~\cite{peebles2023dit}.

The Transformer takes as input a sequence constructed by upsampling each scale's representation to the next scale's resolution:
\begin{equation}
  \mathbf{z} = \big[\,\mathbf{c}_\text{task};\; \text{Up}(\mathbf{a}^{(1)});\; \text{Up}(\mathbf{a}^{(2)});\; \ldots;\; \text{Up}(\mathbf{a}^{(T/2)})\,\big],
  \label{eq:ar_input}
\end{equation}
where $\mathbf{c}_\text{task}$ is a learnable task token and $\text{Up}(\cdot)$ denotes linear upsampling by a factor of 2 along the temporal axis.
At the coarsest scale, $\mathbf{c}_\text{task}$ serves as the sole input; at each subsequent scale, the upsampled output of the previous scale provides the input tokens.
Attention is restricted so that each scale attends only to coarser scales, producing conditioning features:
\begin{equation}
  \hat{\mathbf{s}}^{(i)} = \text{Transformer}(\mathbf{z}, \mathbf{c}_\text{global})^{(i)} \in \R^{i \times D},
  \label{eq:cond_semantics}
\end{equation}
where the superscript $(i)$ selects the output tokens corresponding to scale $i$.
This yields the autoregressive factorization:
\begin{equation}
  p(\mathbf{a}^{(1)}, \ldots, \mathbf{a}^{(T)}) = \prod_{i \in \mathcal{S}} p\big(\mathbf{a}^{(i)} \mid \hat{\mathbf{s}}^{(i)}\big).
  \label{eq:ar_factorization}
\end{equation}
Each conditional distribution $p(\mathbf{a}^{(i)} \mid \hat{\mathbf{s}}^{(i)})$ is modeled by the flow matching module described in Sec.~\ref{subsec:flow_matching}, which implicitly defines the distribution through a learned velocity field.

\subsection{Scale-Wise Flow Matching}
\label{subsec:flow_matching}

At each scale $i$, we generate continuous actions through flow matching~\cite{lipman2023flow} conditioned on $\hat{\mathbf{s}}^{(i)}$.
Given the target action $\mathbf{a}^{(i)} \in \R^{i \times D}$ and a noise sample $\boldsymbol{\epsilon} \sim \mathcal{N}(\mathbf{0}, \mathbf{I})$ of the same shape, we construct an interpolated input at flow time $\tau \in [0, 1]$:
\begin{align}
  \label{eq:interpolant}
  \mathbf{x}^{(i)}_\tau &= (1 - \tau)\, \mathbf{a}^{(i)} + \tau\, \boldsymbol{\epsilon}.
\end{align}
The target velocity field along this straight-line path is:
\begin{align}
  \label{eq:velocity_target}
  \mathbf{v}^{(i)} &= \boldsymbol{\epsilon} - \mathbf{a}^{(i)}.
\end{align}
A shared ActionFlowNet $f_\theta$ predicts the velocity given the noisy input, flow timestep, and the conditioning features:
\begin{align}
  \label{eq:velocity_pred}
  \hat{\mathbf{v}}^{(i)} &= f_\theta\big(\mathbf{x}^{(i)}_\tau,\; \tau,\; \hat{\mathbf{s}}^{(i)}\big).
\end{align}

The ActionFlowNet is a lightweight Transformer with $L_f$ blocks, each employing self-attention with AdaLN modulation conditioned on the sum of a flow timestep embedding and the per-position conditioning from $\hat{\mathbf{s}}^{(i)}$.
The same network is shared across all scales, with the scale naturally handled by varying input sequence lengths.
The conditioning features $\hat{\mathbf{s}}^{(i)}$ are injected via per-timestep AdaLN modulation: unlike the global conditioning used in standard diffusion transformers~\cite{peebles2023dit}, each temporal position in the flow network receives its own modulation parameters derived from the corresponding position in $\hat{\mathbf{s}}^{(i)}$.
This temporally adaptive modulation enables the flow network to generate actions that are coherent both within and across timesteps.

The total training loss aggregates the flow matching MSE across all scales with scale-proportional weighting:
\begin{align}
  \label{eq:total_loss}
  \mathcal{L} &= \frac{1}{|\mathcal{S}|} \sum_{i \in \mathcal{S}} \frac{i}{T} \left\| \hat{\mathbf{v}}^{(i)} - \mathbf{v}^{(i)} \right\|^2,
\end{align}
where the weight $i / T$ places larger weight on finer scales, which capture fine-grained temporal details.
During training, teacher forcing is employed across all scales, where each scale receives the upsampled ground-truth of its preceding scale as input.

\subsection{Training and Inference}
\label{subsec:training_inference}

Unlike prior work that requires a two-stage pipeline of first training an action tokenizer and then training the autoregressive predictor~\cite{gong2024carp}, \methodname trains the entire model end-to-end in a single stage.
The vision encoder, ScaleAR, and ActionFlowNet are jointly trained using the loss in Eq.~\eqref{eq:total_loss}.
This ensures that ScaleAR's conditioning features are tailored to the flow matching objective, without an intermediate tokenization bottleneck.

At inference time, actions are generated autoregressively across scales.
For each scale $i$, ScaleAR generates conditioning features $\hat{\mathbf{s}}^{(i)}$, attending to all coarser scales.
Starting from noise $\mathbf{x}^{(i)}_1 \sim \mathcal{N}(\mathbf{0}, \mathbf{I})$, we integrate the learned velocity field from $\tau=1$ to $\tau=0$ using $N_\text{steps}$ Euler steps:
\begin{equation}
\mathbf{x}^{(i)}_{\tau-\Delta\tau} = \mathbf{x}^{(i)}_\tau - \Delta\tau \cdot f_\theta(\mathbf{x}^{(i)}_\tau, \tau, \hat{\mathbf{s}}^{(i)}), 
  \label{eq:euler}
\end{equation}
yielding the denoised action $\hat{\mathbf{a}}^{(i)} \approx \mathbf{x}^{(i)}_0$.
The result is linearly upsampled to $2i$ timesteps and fed as input tokens for the next scale.

Through the autoregressive generation in Eq.~\eqref{eq:ar_factorization}, the final output $\hat{\mathbf{a}}^{(T)}$ at the finest scale provides the full $T$-step action chunk for open-loop execution.
The total number of neural network forward passes is $|\mathcal{S}| \times (1 + N_\text{steps})$: one Transformer pass plus $N_\text{steps}$ flow steps per scale.

\begin{table}[t]
  \centering
  \caption{Experimental settings.}
  \label{tab:exp_setup}
  \renewcommand{\arraystretch}{1.15}
  \begin{tabular}{ll}
    \toprule
    $D$     & 1024 \\
    $T$     & 8 \\
    $L_e$   & 12 \\
    $L_f$     & 6 \\
    Batch size       &  128 \\
    Optimizer                & AdamW \\
    Learning rate            & $1\times10^{-4}$ \\
    LR schedule              & Cosine \\
    EMA rate                 & 0.9999 \\
    $N_\text{steps}$ & 25 \\
    \bottomrule
  \end{tabular}
\end{table}

\section{Experiments}
\label{sec:experiments}
\begin{figure}[t]
  \centering
  \includegraphics[width=\linewidth]{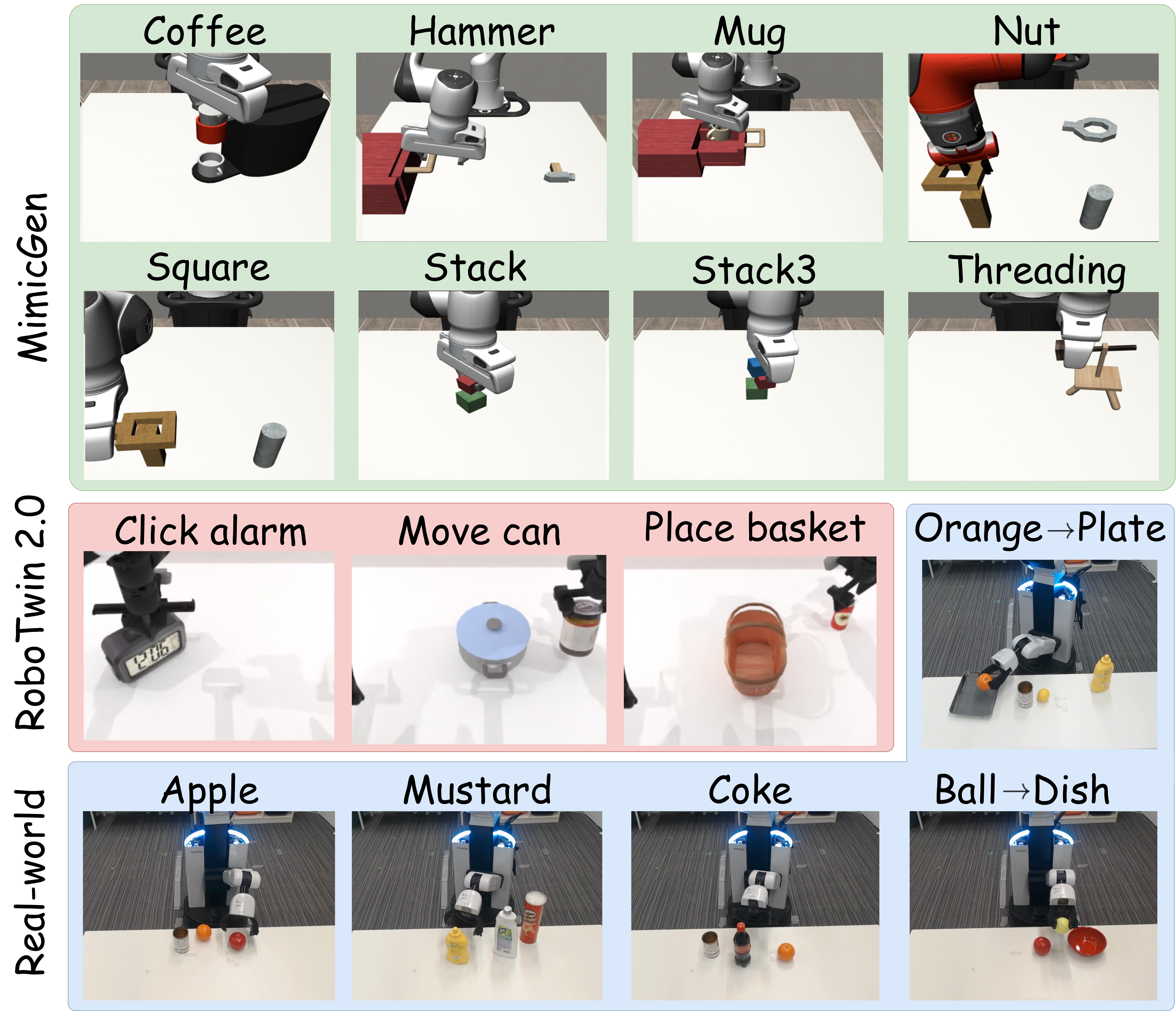}
  \caption{{Task overview across the three evaluation benchmarks.}
Top: 8 single-arm manipulation tasks from MimicGen.
Middle: 3 dual-arm tasks from RoboTwin 2.0.
Bottom: 5 real-world tasks with a mobile manipulator, spanning object grasping, relocation, and target placement.
}
  \label{fig:real_world}
\end{figure}

We evaluate \methodname on three benchmarks spanning diverse environments and embodiments: MimicGen~\cite{mandlekar2023mimicgen} for single-arm multi-task manipulation, RoboTwin 2.0~\cite{chen2025robotwin} for dual-arm manipulation, and real-world experiments with a mobile manipulator. 
Figure~\ref{fig:real_world} shows the experimental setup for each benchmark.

\subsection{Experimental Setup}
\label{subsec:exp_setup}

\begin{table*}[t]
  \centering
\caption{Quantitative results on MimicGen in success rate (\%). Best results in \textbf{bold}.}
\label{tab:mimicgen_results}
  \renewcommand{\arraystretch}{1.15}
  \setlength{\tabcolsep}{4pt}
  \normalsize
  \begin{tabular}{l c c c c c c c c c}
    \toprule
    \textbf{Policy} &
    \textbf{Coffee} & \textbf{Hammer} & \textbf{Mug} & \textbf{Nut} &
    \textbf{Square} & \textbf{Stack} & \textbf{Stack3} & \textbf{Thread} & \textbf{Avg.} \\
    \midrule
    TCD~\cite{liang2024skilldiffuser} &
    77 & 92 & 53 & 44 & 63 & 95 & 62 & 56 & 68 \\
    SDP~\cite{wang2024sdp} &
    82 & \bf{100} & 62 & 54 & 82 & 96 & 80 & 70 & 78 \\
    CARP~\cite{gong2024carp} &
    86 & 98 & 74 & 78 & \bf{90} & \bf{100} & \bf{82} & 70 & 85 \\
    \midrule
    \methodname (Ours) &
    \bf{100} & \bf{100} & \bf{76} & \bf{80} & 82 & 98 & 76 & \bf{90} & \bf{88} \\
    \bottomrule
  \end{tabular}
\end{table*}

 \begin{table}[t]
\caption{Quantitative results on RoboTwin benchmark. Success rates (\%). Best results in \textbf{bold}. ``Click alarm'', ``Move can'', and ``Place basket'' denote ``click alarmclock'', ``move can pot'', and ``place can basket'', respectively.} 
  \label{tab:robotwin_results}
  \footnotesize
  \centering
  \renewcommand{\arraystretch}{1.15}
  \setlength{\tabcolsep}{3pt}
  \begin{tabular}{l c c c}
    \toprule
    \textbf{Policy} &
    \textbf{Click alarm} &
    \textbf{Move can} &
    \textbf{Place basket} \\
    \midrule
    ACT~\cite{zhao2023act} &
    32 & 22 & 1  \\
    DP~\cite{chi2023diffusionpolicy} &
    61 & 32 & 18  \\
    \midrule
    \methodname (Ours) &
    \bf{69} & \bf{42} & \bf{39}  \\
    \bottomrule
  \end{tabular}
\end{table}

\begin{table}[t]
  \centering
  \caption{Quantitative results on real-world experiments.
  Success rates (\%) over 12 trials per task. Best results in \textbf{bold}.}
  \label{tab:real_world_results}
  \footnotesize
  \renewcommand{\arraystretch}{1.15}
  \setlength{\tabcolsep}{3pt}
    \begin{tabular}{@{}l c cc cc@{}}
    \toprule
    & \textit{Grasp} & \multicolumn{2}{c}{\textit{Relocation}} & \multicolumn{2}{c}{\textit{Target Placement}} \\
    \cmidrule(lr){2-2} \cmidrule(lr){3-4} \cmidrule(lr){5-6}
    \textbf{Policy} &
    \textbf{Apple} &
    \textbf{Mustard} &
    \textbf{Coke} &
    \textbf{Ball$\rightarrow$Dish} &
    \textbf{Orange$\rightarrow$Plate} \\
    \midrule
    ACT~\cite{zhao2023act} &
    50.0 & 33.3 & 8.3 & 16.7 & 8.3 \\
    DP~\cite{chi2023diffusionpolicy} &
    66.7 & 58.3 & 66.7 & 25.0 & 16.7 \\
    CARP~\cite{gong2024carp} &
    66.7 & 58.3 & 58.3 & 33.3 & 25.0 \\
    \midrule
    \methodname (Ours) &
    \textbf{75.0} & \textbf{83.3} & \textbf{75.0} & \textbf{58.3} & \textbf{41.7} \\
    \bottomrule
  \end{tabular}
\end{table}

\textbf{MimicGen.}
We evaluate on 8 manipulation tasks from MimicGen~\cite{mandlekar2023mimicgen}: Coffee, Hammer Cleanup, Mug Cleanup, Nut Assembly, Square, Stack, Stack Three, and Threading.
Each task has 1K--10K demonstrations with image observations from a head camera and a wrist camera, along with proprioceptive robot states.
Models are trained on all 8 tasks jointly with task conditioning.

We compared \methodname with three baselines: task-conditioned diffusion (TCD)~\cite{liang2024skilldiffuser}, a standard diffusion-based multi-task policy; Sparse Diffusion Policy (SDP)~\cite{wang2024sdp}, a transformer-based diffusion policy that leverages mixture of experts; and CARP~\cite{gong2024carp}, an autoregressive policy that tokenizes actions via VQ-VAE~\cite{vqvae}.
All baselines were trained with visual inputs following their official implementations and settings.
We followed the evaluation setup in~\cite{gong2024carp}, where only strict task completion was counted as a success.

\textbf{RoboTwin 2.0.}
To assess dual-arm manipulation capability, we use the RoboTwin benchmark~\cite{chen2025robotwin}, a simulation platform featuring bimanual 6-DoF arms with parallel-jaw grippers (14-DoF total).
The policy receives image observations from a fixed head camera and two wrist-mounted cameras, along with proprioceptive end-effector states.
We select 3 representative tasks: ``click alarmclock'', ``move can pot'', and ``place can basket'', each providing 50 training demonstrations and 100 evaluation rollouts.
We compare \methodname with two baselines: ACT~\cite{zhao2023act} and Diffusion Policy (DP)~\cite{chi2023diffusionpolicy}, two widely adopted visuomotor policies for imitation learning.

\textbf{Real-world experiments.}
To validate that our approach transfers beyond simulated tabletop settings, we deploy \methodname on a physical mobile manipulation platform with a different embodiment from the simulation benchmarks.
Specifically, we use Toyota Motor Corporation's Human Support Robot (HSR), a mobile manipulator with 11-DoF that has served as the standard platform for RoboCup@Home since 2017~\cite{iocchi15aij}.
The policy receives image observations from a head-mounted camera and a hand camera, along with proprioceptive robot states.

We design 5 tasks spanning three categories of increasing difficulty: \textit{object grasping}, where the robot picks up a target object (Apple); \textit{object relocation}, where the robot grasps an object and moves it to a designated area (Mustard, Coke); and \textit{target placement}, where the robot grasps an object and precisely places it onto a specific container (Ball$\rightarrow$Dish, Orange$\rightarrow$Plate).
Objects are chosen to cover diverse geometries, including spherical, cylindrical, and irregular shapes.
We collect 50 demonstration trajectories per task and evaluate over 12 trials each with randomized object placements and robot initial poses.
The demonstrations were collected using a leader-follower system~\cite{takanami2025arxiv} at a sampling rate of 10Hz.
We compare \methodname with three baselines: DP~\cite{chi2023diffusionpolicy}, ACT~\cite{zhao2023act}, and CARP~\cite{gong2024carp}.

\subsection{Implementation details}
Table~\ref{tab:exp_setup} shows the experimental setup for \methodname.
We used 8 NVIDIA H200 SXM GPUs (141GB VRAM) for training and a single GeForce RTX 4090 (VRAM 24GB) for inference.
All actions were normalized to $[-1, 1]$ using quantile bounds.
We report per-task success rates for all benchmarks.

\begin{figure}[t]
  \centering
  \includegraphics[width=\linewidth]{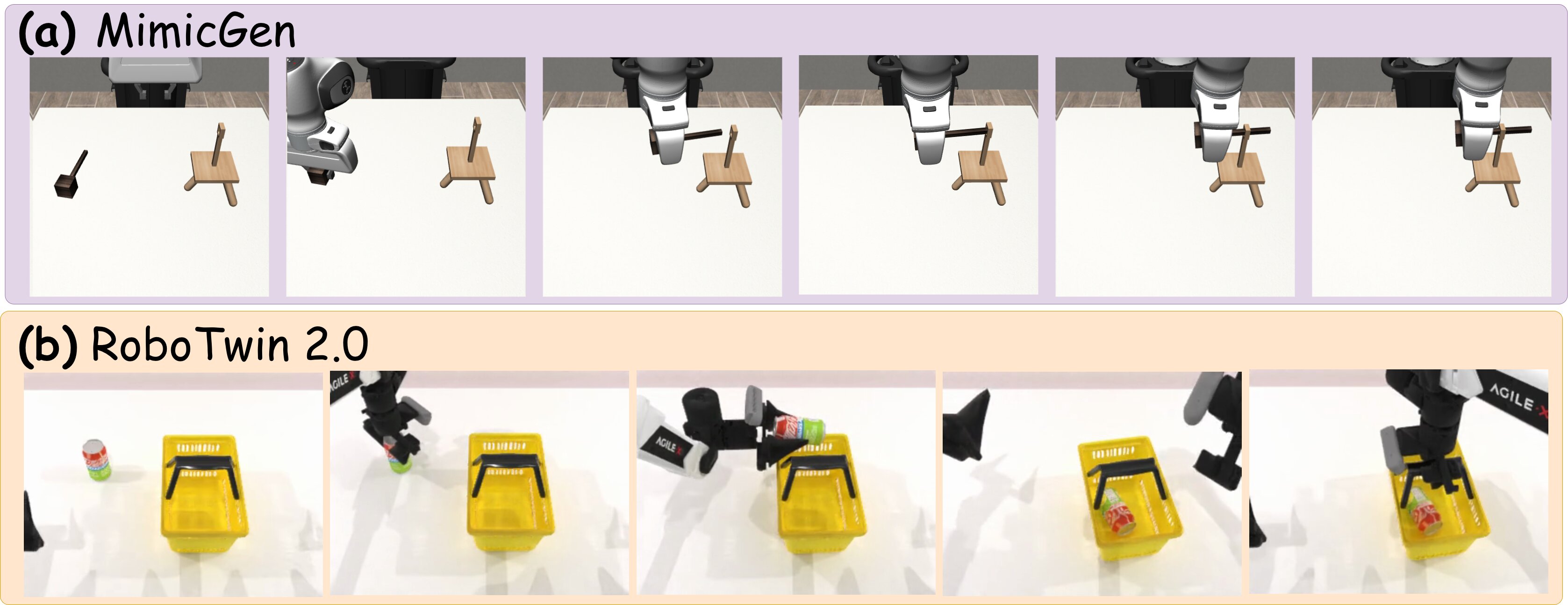}
  \caption{Qualitative results of \methodname on representative tasks from each simulation benchmark. (a)~\textit{Threading} from MimicGen. (b)~\textit{Place can basket} from RoboTwin 2.0.
  }
  \label{fig:qualitative_results}
  \vspace{-2mm}
\end{figure}

\begin{figure*}[t]
  \centering
  \includegraphics[width=\linewidth]{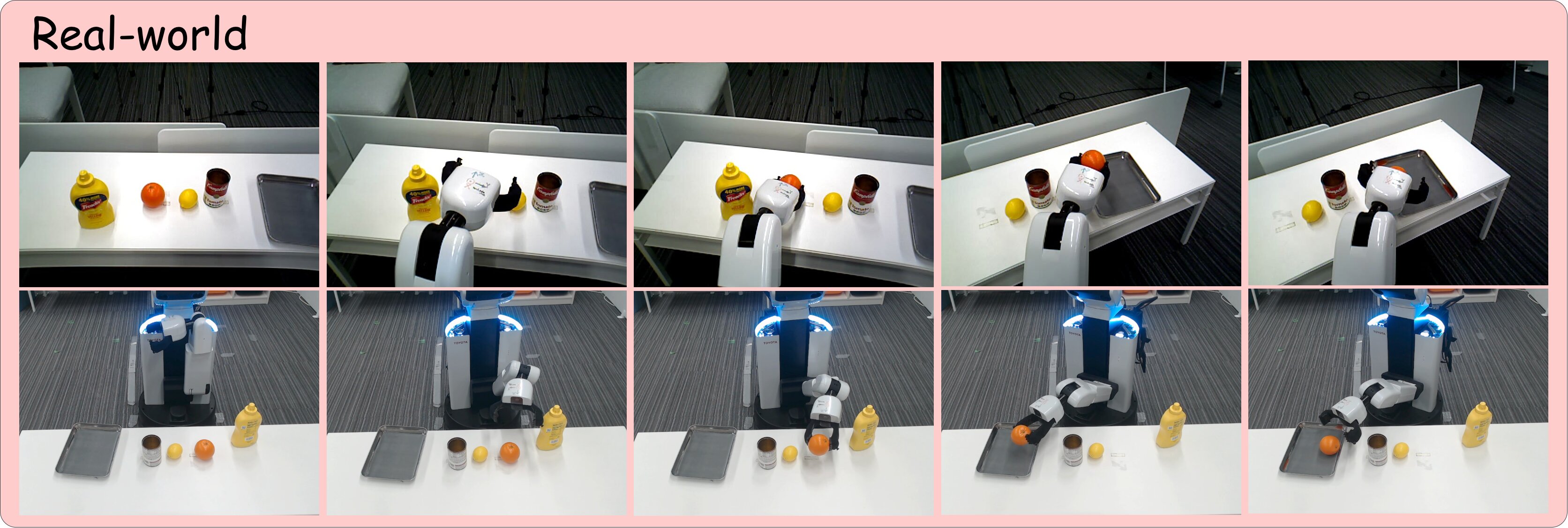}
  \caption{Qualitative results of \methodname on the \textit{Orange$\rightarrow$Plate} task from the real-world experiments. Note that the third-person view are not included in the observation of the model.
  }
  \label{fig:qualitative_results_hsr}
  \vspace{-2mm}
\end{figure*}

\begin{figure*}[t]
  \centering
  \includegraphics[width=\linewidth]{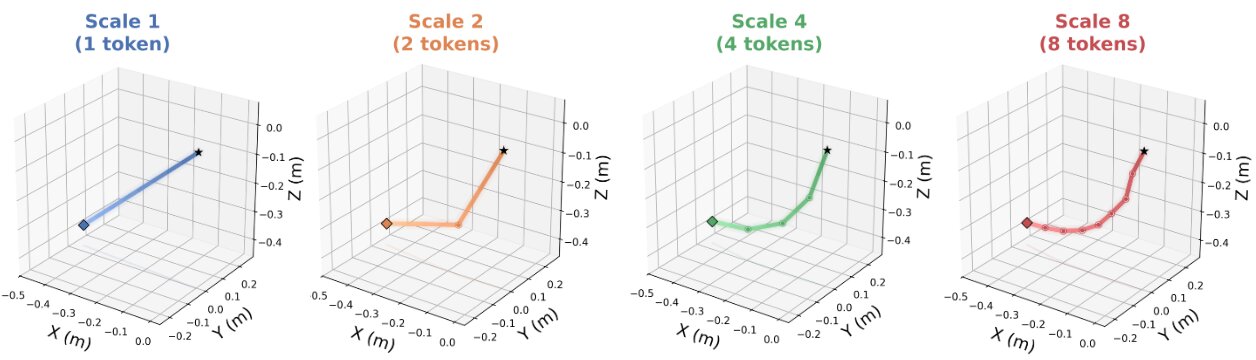}
  \caption{{Visualization of coarse-to-fine action generation.}
  Each panel shows the predicted cumulative end-effector displacement trajectory in Cartesian task space (position
  components only) at a different scale level on the Mimicgen benchmark. 
  $\circ$,$\star$, and $\diamond$ indicate token boundaries, trajectory start point and the end point, respectively. 
  }
  \label{fig:trajectory_viz}
\end{figure*}

\subsection{Quantitative Results}
Tables~\ref{tab:mimicgen_results} and ~\ref{tab:robotwin_results} show the quantitative results on the MimicGen and RoboTwin benchmarks, respectively.
As shown in these tables, \methodname consistently outperformed all baselines on both benchmarks.
On MimicGen, \methodname achieved an average success rate of 88\%, surpassing the strongest baseline CARP at 85\% while outperforming diffusion-based methods TCD and SDP by margins of 20\% and 10\%, respectively.
Notably, \methodname achieved perfect scores on \textit{Coffee} and \textit{Hammer}, and attained the largest improvement on \textit{Threading} with 90\% compared to 70\% for both SDP and CARP.
Threading is a fine-grained manipulation task that demands precise action trajectories, suggesting that flow matching better captures detailed motion patterns than both diffusion denoising and VQ-VAE tokenization.

On RoboTwin, where policies need to coordinate bimanual 7-DoF arms, \methodname showed even larger gains over the baselines.
The improvement was most notable on \textit{place can basket}, a task requiring coordinated dual-arm grasping and placement, where \methodname achieved 39\% compared to 18\% for Diffusion Policy and just 1\% for ACT.
This demonstrated that our tokenization-free formulation scales effectively to higher-dimensional action spaces.

Table~\ref{tab:real_world_results} presents the results of the real-world experiments.
As shown in Table~\ref{tab:real_world_results}, \methodname consistently outperformed all baselines across every task.
The performance gap widened further in the more challenging target placement tasks, where precise positioning is required.
\methodname achieved 58.3\% and 41.7\% on Ball$\rightarrow$Dish and Orange$\rightarrow$Plate, respectively, compared to 33.3\% and 25.0\% for the strongest baseline.
These results confirmed that the advantages observed in simulation transferred to real-world settings, where perception noise and physical contact dynamics introduce additional challenges beyond those present in controlled environments.

\subsection{Qualitative Results}
\label{subsec:qual_results}
Figures~\ref{fig:qualitative_results} and \ref{fig:qualitative_results_hsr} presents successful rollouts of \methodname on representative tasks from each of the three evaluation settings.
Figure~\ref{fig:qualitative_results}(a) shows the \textit{Threading} task from MimicGen, where \methodname generated smooth trajectories to align the needle with the narrow hole and insert it.
This illustrated the advantage of continuous flow matching in capturing fine-grained motion details.
In Figure~\ref{fig:qualitative_results}(b), for the \textit{Place can basket} task in the RoboTwin benchmark, the left arm first grasped the can and placed it into the basket, after which the right arm picked up the basket.
This sequential bimanual coordination highlighted that our tokenization-free formulation scales naturally to high-dimensional action spaces requiring inter-arm dependency.
As depicted in Figure~\ref{fig:qualitative_results_hsr}, in the real-world \textit{Orange$\rightarrow$Plate} experiment, \methodname reliably grasped the orange and placed it at the target location on the plate.
Notably, \methodname produced smooth and consistent placement trajectories without abrupt corrections, indicating that the continuous flow matching formulation generates temporally coherent actions even under real-world sensory noise.

Figure~\ref{fig:trajectory_viz} visualizes the coarse-to-fine action generation process of \methodname across different scale levels.
At Scale~1, a single token captured the overall direction of the trajectory, providing a rough global motion.
As the scale increased, additional tokens progressively refined the trajectory, introducing finer temporal detail and curvature.
By Scale~8, the predicted trajectory closely followed the detailed motion required to complete the task.
This confirmed that our multi-scale formulation enables \methodname to first establish a high-level motion structure and then gradually fill in the details, mirroring the coarse-to-fine nature of flow matching.

\begin{table}[t]
  \centering
  \caption{Sensitivity analysis on temporal downsampling scales. Average success rate (\%) on MimicGen. Best result in \textbf{bold}.}
  \label{tab:ablation_scales}
  \begin{tabular}{@{}l c@{}}
    \toprule
    \textbf{Scale configuration} & \textbf{Avg. SR} \\
    \midrule
    $\{1, 8\}$          & 84 \\
    $\{1, 4, 8\}$        & 86 \\
    $\{1, 2, 4, 8, 16\}$    & 85 \\
    \midrule
    $\{1, 2, 4, 8\}$ & \textbf{88} \\
    \bottomrule
  \end{tabular}
\end{table}
\subsection{Sensitivity Analysis}
\label{subsec:ablations}
We conducted sensitivity analysis on the MimicGen benchmark to investigate the effect of the temporal scale configuration by varying the number of scales used in \methodname.
Table~\ref{tab:ablation_scales} summarizes the results.
Using only two scales $\{1, 8\}$ yielded an average success rate of 84\%, as the large stride gap between scales forced the model to bridge coarse and fine actions in a single step, losing intermediate temporal structure.
Adding an intermediate scale $\{1, 4, 8\}$ improved performance to 86\%, confirming that a gradual coarse-to-fine hierarchy helped the model capture progressively finer motion details.
Our default configuration with four scales $\{1, 2, 4, 8\}$ achieved the best result at 88\%, providing a well-balanced temporal hierarchy where each scale refined the previous one at a manageable resolution gap.
However, further increasing to five scales $\{1, 2, 4, 8, 16\}$ slightly degraded performance to 85\%, suggesting that excessively aggressive downsampling at the coarsest level generated overly compressed representations that propagated errors through the hierarchy.
Nonetheless, all configurations achieved strong success rates, demonstrating that \methodname is robust to the choice of temporal scale configuration.

\section{Conclusion}
\label{sec:conclusion}

We introduced \methodname, a tokenization-free coarse-to-fine autoregressive policy for visuomotor robot manipulation.
By recognizing that the discrete tokenization paradigm borrowed from image generation is unnecessary for low-dimensional, dense action spaces, we replaced learned VQ-VAEs with simple temporal averaging and cross-entropy prediction with continuous flow matching.
This yields a single-stage training pipeline that eliminates quantization error, preserves the Euclidean geometry of actions, and naturally handles multimodal distributions.
Experiments on MimicGen, RoboTwin 2.0, and real-world tasks demonstrate that \methodname consistently outperforms existing diffusion-based and tokenization-based autoregressive policies.

{\small
  \bibliographystyle{IEEEtran}
  \bibliography{reference}
}

\end{document}